\documentclass{article}

\usepackage[nonatbib, final]{nips_2017}
\usepackage{subfig}
\usepackage{enumitem} 
\usepackage{algorithm}
\usepackage{algpseudocode}

\usepackage[utf8]{inputenc} % allow utf-8 input
\usepackage[T1]{fontenc}    % use 8-bit T1 fonts
\usepackage[pdfborder={0 0 0}]{hyperref}       % hyperlinks
\usepackage{url}            % simple URL typesetting
\usepackage{booktabs}       % professional-quality tables
\usepackage{amsfonts}       % blackboard math symbols
\usepackage{nicefrac}       % compact symbols for 1/2, etc.
\usepackage{microtype}      % microtypography
\usepackage{amsmath}		% math library
\usepackage{graphicx}
\usepackage{caption}
\usepackage{url}  
\usepackage{float}  
\newtheorem{theorem}{Theorem}
\begin{document}
\title{Global overview of Imitation Learning}

% The \author macro works with any number of authors. There are two
% commands used to separate the names and addresses of multiple
% authors: \And and \AND.
%
% Using \And between authors leaves it to LaTeX to determine where to
% break the lines. Using \AND forces a line break at that point. So,
% if LaTeX puts 3 of 4 authors names on the first line, and the last
% on the second line, try using \AND instead of \And before the third
% author name.

\author{
  Alexandre Attia\\
  MVA \\
  ENS Paris-Saclay\\
  \texttt{alexandre.attia@ens-paris-saclay.fr} \\\And
  Sharone Dayan   \\
  MVA \\
  ENS Paris-Saclay\\
  \texttt{sharone.dayan@ens-paris-saclay.fr} \\
}

\maketitle

\begin{abstract} 
  Imitation Learning is a sequential task where the learner tries to mimic an expert's action in order to achieve the best perfomance. Several algorithms have been proposed recently for this task. In this project, we aim at proposing a wide review of these algorithms, presenting their main features and comparing them on their performance and their regret bounds.
\end{abstract}

\section{Introduction}
The principle behind Imitation Learning is to act and exhibit a human behavior by implicitly giving to a learner prior information about the world. In Imitation Learning tasks, the agent seeks the best way to use a training set (input-output pair) demonstrated by an expert in order to learn a policy and achieve an action as similar as possible as the expert's one. Imitation is often needed to automate actions when the agent is human and it is too expensive to run its actions in real-time. Apprenticeship learning[1], on the contrary,  executes pure greedy/exploitative policies and use all (state/action) trajectories to learn a near-optimal policy using Reinforcement Learning approaches. It requires difficult maneuvers and it is nearly impossible to recover from unobserved states. Imitation learning can often deal with those unexplored states so it offers a more reliable framework for many tasks such as self-driving cars.
%Iterative imitation learning algorithms : the agent actively queries the expert for demonstrations in states that it sees when executing its current policy.\\
We will first setup the notations and the general framework, then we will present some of the main Imitation Learning algorithms and their guarantees of convergence. Finally, we will focus on experimental results of the DAgger approach on a real-world application.
%and test it on real system.
%exploration/exploitation dilemma

\section{Problem setup}
Let us introduce imitation learning in the framework of Markov Decision Processes (MDP). \\A MDP is defined by the tuple $<S,A,B,R,I>$ with : $S$ the set of states, $A$ the finite set of actions, $B(s,a,s')$ the transition function, $R(s,a)$ the reward ($\in$ [0,1]) of performing action $a$ in state $s$ and $I$ the initial state distribution. We denote $N$ as being the number of epochs. \\The policy we make use of can either be stationary \textit{(Markovian)} $\pi=(\pi,...,\pi)$ or non-stationary  $\pi=(\pi_0,...,\pi_T)$ with $T$ being the time horizon. It indicates the action to take in state $s$ and at time $t$. We denote the deterministic expert policy as $\pi^*$ and use the following notations : 
\begin{itemize}
\itemsep0em
    \item $J(\pi)$ the expected total reward of trajectories starting with the initial state $I$
    \item $d_{\pi} =\frac{1}{T}\sum_{t=1}^T d_{\pi}^t$ the empirical mean of state distribution induced over each time step
    \item $C_{\pi}(s) = \mathbb{E}_{a \sim \pi(s)}[R(s,a)]$ the total reward in a $T$-step trajectory
    \item $l(s, \pi)$ the observed surrogate loss 
\end{itemize}
With the previous notations, we obtain the following quantity that we aim at maximizing : $$J(\pi) = T \mathbb{E}_{s \sim d_\pi}[C_{\pi}(s)]$$ 
There exists two types of settings that can be applied: the passive setting where the learner is provided with a training set of optimal policy's execution full trajectories and the active setting where the learner is allowed to pose action queries to an expert that returns the desired action for a specific time. %We also denotes $N$ the number of epochs. 
%The goal is thus to maximize the expectation of the total reward., i.e minimize the regret $J(\pi)-J(\pi^*)$. be the total cost (also called cost-to-go). 
%and the state-action pairs are then passed to an i.i.d. supervised learning algorithm

\section{State-of-the-art algorithms and their guarantee of convergence}
Learning from Demonstration (LfD) is a practical framework for learning complex behaviour policies from demonstration trajectories produced by an expert, even if there are very few of them or if they are inaccurate. We list and compare here some of the most used algorithms for imitation learning where we have drawn illustrations of the models on a self-driving car to spot the differences between the various algorithms. Some of the theoretical proofs and intuitions of the theorems below are in Appendix.

\subsection{Supervised learning}
The first approach to tackle imitation learning is supervised learning by classification. We have a set of training trajectories (stationary policy) achieved by an expert where a single trajectory $t$ consists of a sequence of observations and a sequence of actions executed by an expert. The motivation behind imitation learning with supervised learning is to teach a classifier that attempts to mimic the expert's action based on the observations at that time. 

It is a passive approach where the objective is to learn a target policy by passively observing full execution trajectories. The expert acts only before solving the learning objective which is to train a policy over the states encountered by the expert. Also, we need to make the assumption that actions in the expert trajectories are independent identically distributed (i.i.d).

There exists an upper bound on the loss suffered by the Supervised Imitation Learning algorithm as a function of the quality of the expert and the error rate of the learned classifier. Let $\epsilon$ be the error rate of the underlying classifier, T the horizon and $\pi$ the learned policy, we have a quadratic cost. \begin{theorem}
Let denote $\epsilon = \mathbb{E}_{s \sim d_{\pi^*}}[l(s, \pi, \pi^*)]$, then there exists $\pi \in \pi_{1:N}$ such that \\ $J(\pi) \leq J(\pi^*)+ T^2\epsilon$
\end{theorem}

%The intuition behind the quadratical 

The main issue with this supervised learning approach to imitation learning is that it cannot learn to recover from failures. Indeed, supposing that the model has deviated from the optimal trajectory at one time step, it will not be able to get back to states seen by the expert and hence, it will generate a cascade of errors. We conclude that this naive algorithm fails to generalize to unseen situations. The next approaches rectify this behaviour. %On the concrete example of self-driving cars, if the steering has gone the wrong direction

\subsection{Forward Training}

The forward training algorithm was introduced by Ross and Bagnell (2010)[2] and it trains one policy $\pi_t$ at each time step $t$ over $T$ (non-stationary policy), i.e at each $t$, the machine learns a policy $\pi_t$ to mimic the expert policy $\pi^*$ on the states induced by the previous learned policies $\pi_1, ..., \pi_{t-1}$. This iterative training is described in Algorithm \ref{alg:ft}.\\
Let $u$ be be the maximal increase in the expected total cost from any probable state, when changing only the policy. For this algorithm, we have a guaranteed performance with a near-linear regret. 
\begin{theorem}
Let denote $\epsilon = \mathbb{E}_{s \sim d_{\pi^*}}[l(s, \pi, \pi^*)]$, then there exists $u$ and $\pi \in \pi_{1:N}$ such that \\ \textbf{$J(\pi) \leq J(\pi^*)+ O(u T \epsilon)$} 
\end{theorem}
In the worst case, we have the same convergence as for classical supervised learning but in general, the convergence is sublinear and the experts policies succeed in recovering the mistakes of the model policy. Thus, the Forward Training algorithm should perform better than the previous one. 

\begin{algorithm}[H]
\caption{Forward Training}\label{alg:ft}
\begin{algorithmic}[1]
\State $\text{Initialize } \pi^0_1, ..., \pi^0_T \text{ to query and execute } \pi^*$
\For {$i=1...T$}
\State $\text{Sample T-step trajectories by following} \pi^{i-1}$
\State $\text{Get dataset } D=\{(s_i, \pi^*(s_i))\} \text{ of (states, action) taken by expert at step } i$ 
\State $\text{Train classifier } \pi_i^i = argmin_{\pi \in \Pi}\mathbb{E}_{s \sim D}(e_{\pi}(s))$
\State $\pi_j^i = \pi_j^{i-1} \ \forall j \neq i $
\EndFor \\
\Return $\pi^T_1, ..., \pi^T_T$
\end{algorithmic}
\end{algorithm}

However, one major weakness of the presented approach is that it needs to iterate over all the T periods, where the time horizon T can be quite large or even undefined. Thus, taken that the policy is non-stationary, the algorithm becomes impracticable in most real-world applications (T large or undefined). Some of the next algorithms overcome this issue.

\subsection{Search-based Structured Prediction (SEARN)}
 The idea behind SEARN introduced by Daumé III et al. (2009) [3] is that instead of learning some sort of global model and then searching (as it is the standard), it will simply learn a classifier to make each of the decisions of the search optimally. The algorithm starts by following the experts action at every step. Iteratively, it collects the demonstrations and make use of them to train a new policy. It compiles new episodes by taking actions according to a mixture of all previously trained policies, as well as the experts actions. Finally, over time, it learns to follow its mixture of policies and stops relying on the expert to decide which actions to take.
 
 In short, this algorithm attempts to learn a classifier that will walk us through the search space. It operates by maintaining a current policy and attempts to use it in order to generate new training data on which to learn a new policy (new classifier). When a new classifier is learned, we interpolate it with the old classifier. This iterative scheme  is described in Algorithm \ref{alg:searn}. \\
  We can bound the cost as explained below in Theorem~\ref{thm_searn}.

\begin{theorem}
Using $\alpha$ in $O(T^{-3})$ and $N$ in $O(T^3 lnT)$ then, there exists $\pi \in \pi_{1:N}$ and $\mathbb{A}_1$ such as
$ J(\pi) < J(\pi^*)+ O(\mathbb{A}_1Tln T + ln T) $
\label{thm_searn}
\end{theorem}

%\url{http://www.umiacs.umd.edu/~hal/searn/}\\
%SEARN is mostly used for Natural Language Processing (NLP). 
%This method allows the learning to take into account the fact that we will be running a search algorithm. Instead of training based on the true path through the search space, train on the path that our classifier actually takes in practice. This means that the classifier will be trained based on the data that it will actually expect to see.SEARN operates by maintaining a current policy and attempts to use to this generate new training data on which to learn a new policy (new classifier). The current policy is initialized to the optimal policy. When a new classifier is learned, we interpolate it with the old classifier.\\

\begin{algorithm}
\caption{SEARN}\label{alg:searn}
\begin{algorithmic}[1]
\State $\text{Initialize } \pi \gets \pi^* \text{ (optimal policy)}$
\While {$\pi \text{ has a significant dependence on } \pi^*$}
\State $\text{Initialize the set of examples } S \gets \emptyset$
\For {$(x,y) \in \text{ Structured Examples}$}
\State $\text{Compute predictions under the current policy to produce full output } \pi$
\For {$t=1...T_x$} \textit{ (create a single cost-sensitive example)}
\State $\text{Computer features } \Phi = \Phi(s_t) \text{ for state } s_t = (x, y_1, ..., y_t)\text{, initialize a cost vector } c$
\For {each possible action $a$}
\State $\text{Let the cost } l_a \text{ for example } (x,c) \text{ at state } s \text{ be } l_\pi(c,s,a)$
\EndFor
\State $\text{Add cost-sensitive example} (\Phi, l) \text{ to } S$
\EndFor
\EndFor
\State $\text{Learn a new classifier on } S : \pi' \gets A(S)$
\State $\text{Interpolate : } \pi \gets \beta \pi' + (1-\beta)\pi'$
\EndWhile \\
\Return $\pi \text{ without }\pi^*$
\end{algorithmic}
\end{algorithm}

However, this Search-based structured prediction can be overly optimistic and is challenging in practice mainly due to its initialization which is different from the optimal policy. Below, we will detail other approaches that overcome this issue.

\subsection{Stochastic Mixing Iterative Learning (SMILe)}

The SMILe algorithm was also introduced by Ross and Bagnell (2010)[2] to correct some of the inconvenient issues of the forward training algorithm. It is a stochastic mixing algorithm based on SEARN that uses its benefits with a
substantially simpler implementation and less demanding interaction with an expert. It trains a stochastic stationary policy over several iterations and then makes use of a “geometric” stochastic mixing of the policies trained.

Concretely, we start with a policy $\pi_0$ that follows exactly the expert's actions. At each iteration $i$, we train a policy $\pi_i$ to mimic the expert under the trajectories induced by the previous policy $\pi_{i-1}$. Then, we add the new trained policy to the previous mix of policies with a geometric discount factor $\alpha (1-\alpha)^{i-1}$. So, the new policy $\pi_i$  is a mix of $i$ policies, with the probability of using the expert's action as $(1-\alpha)^{i}$. The SMILe algorithm is described in Algorithm~\ref{alg:smile}.\\Selecting $\alpha = O(\frac{1}{T^2})$ and $ N = O(T^2 log (T))$ guarantees near-linear regret (with $u$ the policy disadvantage) as for the forward training algorithm.
\begin{theorem}
Let $\Tilde{\epsilon}$ denote a variable depending on $\mathbb{E}_{s \sim d_{\pi^{i-1}}}(e(s, \widehat{\pi}^{*i}))$. Then there exists $\pi \in \pi_{1:N}$ and $\mathbb{A}_k(\pi,\pi^*) = \Bar{J}_k^{\pi}(\pi')-J(\pi)$ (the k-th order policy disadvantage of $\pi'$ with respect to $\pi$) such that  $J(\pi^N) \leq J(\pi^*) + O(T(\mathbb{A}_1+\Tilde{\epsilon}) + 1) $
\end{theorem}

\begin{algorithm}
\caption{SMILe}\label{alg:smile}
\begin{algorithmic}[1]
\State $\text{Initialize } \pi^0 \gets \pi^* \text{ to query and execute expert}$
\For {$i=1...N$}
\State $\text{Execute } \pi^{i-1} \text{ to get } D=\{(s, \pi^*(s))\}$
\State $\text{Train classifier } \widehat{\pi}^{*i} = argmin_{\pi \in \Pi}\mathbb{E}_{s \sim D}(e_{\pi}(s))$
\State $\pi^i = (1-\alpha)^i\pi^* + \alpha \sum_{j=1}^i  (1-\alpha)^{j-1}\widehat{\pi}^{*j}$
\EndFor 
\State $\text{Remove expert queries } \widetilde{\pi}^N = \frac{\pi^N (1-\alpha)^N \pi^*}{1-(1-\alpha)^N}$\\
\Return $\widetilde{\pi}^N$
\end{algorithmic}
\end{algorithm}
%doublefig
\begin{figure}[!htb]
\centering
\subfloat[Forward training]         {\includegraphics[width=.25\linewidth]{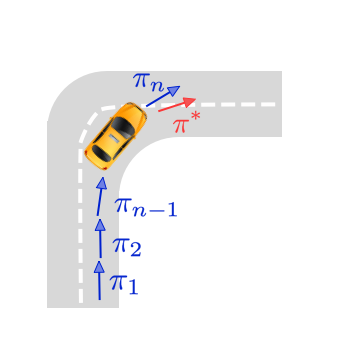}}\hspace{0.5cm}
\subfloat[SMILe]  
  {\includegraphics[width=.25\linewidth]{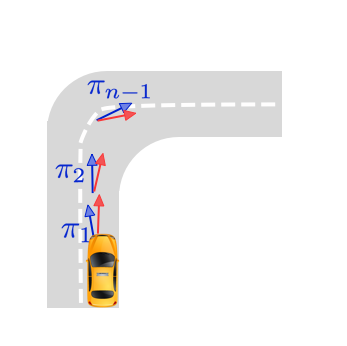}}
\caption{Comparison of Forward training and SMILe algorithms on self-driving cars}
\label{fig:dag}
\end{figure}

The main advantage of this approach is that we can interrupt the process at any time in order to not take into account a too large or undefined time horizon. Unfortunately, due to its stochastic policy, the model is not stable.

\subsection{Reduction-based Active Imitation Learning (RAIL)}

The principle behind RAIL introduced by  Ross et al. (2011) [4] is to perform a sequence of $T$ calls to an independent identically distributed (i.i.d) active learner $L_a$. We note that it is likely to find a useful stationary policy well before all T calls are issued, which palliates the drawbacks of forward training. Indeed, the active learner is able to ask queries across a range of time point and we might expect policies learned in earlier iterations to achieve non-trivial performance throughout the entire horizon.

Concretely, RAIL iterates for $T$ iterations with the notable difference that on each iteration, it learns a new stationary policy that can be applied across all time steps. Iteration $t + 1$ of the model learns a new policy that achieves a low error rate at predicting the expert’s actions with respect to the state distribution of the previous policy. The RAIL algorithm is described in Algorithm~\ref{alg:rail}.

%Reductions from active imitation learning to active i.i.d. learning for the cases of deterministic non-stationary and stationary policies.

\begin{algorithm}
\caption{RAIL}\label{alg:rail}
\begin{algorithmic}[1]
\State $\text{Initialize } \hat{\pi}^0 \text{an arbitrary policy, possibly based on prior knowledge or existing data}$
\For {$t=1...T$}
\State $\hat{\pi}^t = L_a(\epsilon, \delta/t, d_{\hat{\pi}^{t-1}})$
\EndFor \\
\Return $\hat{\pi}^T$
\end{algorithmic}
\end{algorithm}

RAIL is an idealized algorithm intended for analysis which achieves the theoretical goals. However, it has a number of inefficiencies from a practical perspective mainly because of  the unlabeled state distributions
used at early iterations that can be quite different from $d_{\pi^*}$.

\subsection{Dataset Aggregation (DAgger)}

\subsubsection{DAgger } 
Ross and Bagnell proposed, in 2010, the DAgger[5] algorithm to also solves the Learning from Demonstration problems. DAgger is an iterative policy training algorithm via a reduction to online learning. At each iteration, we retrain the main classifier on all states ever encountered by the learner. The main advantage of DAgger is that the expert teaches the learner how to recover from past mistakes. It's an active method (we need access to expert themselves) based on \textit{Follow-The-Leader} algorithm (each iteration is one online-learning example).

We start with a first policy $\pi_0$ fully taught by the expert then, we run $\pi_0$ and see what configurations the learner visits. We generate a new dataset that contains information about how to recover from the errors of $\pi_0$. Because we want to have information from both $\pi_0$ and $\pi_1$, we trained $\pi_1$ on the union of the initial expert-only trajectories together with new generated trajectories. We repeat it at each iteration. We choose the best policy on the validation test.

\begin{theorem} Let $\epsilon_N = \min_{\pi \in \Pi} \frac{1}{N}\sum_{i=1}^N \mathbb{E}_{s \sim d_{\pi_i}}[l(s,\pi)]$ be the true loss of the best policy, then if $N = O(\frac{T}{\epsilon})$ there exists $\pi \in \pi_{1:N}$ and $u$ such as $J(\pi) \leq J(\pi^*) + u T \epsilon_N + O(1)$
\end{theorem}

The main algorithmic difference between SEARN[3] and DAgger is in the learning of the classifiers in each iteration and in combining them into a policy. DAgger can combine the training signal obtained from all iterations contrary to SEARN wich only train on iteration $i$ ie with no aggregate dataset. SEARN was the first practical method, followed by DAgger. DAgger works for both complex and simple problems, it improves the more data is collected but only needs few iterations to work. So it can be useful for many applications such as handwritten recognition or autonomous driving.\\

\begin{algorithm}
\caption{DAgger and DAgger by coaching algorithms}\label{alg:dagger}
\begin{algorithmic}[1]
\State $\text{Initialize } D \gets \emptyset \text{, } \pi_1 \gets \pi^*$
\For {$i = 1:N$}:
\State $\text{Sample T-step trajectories using } \pi_i$
\If {coaching} 
\State $\pi_{target} = \widetilde{\pi_i} \textit{ (Hope Action)}$
\Else
\State $\pi_{target} = \pi_i^* \textit{ (Expert Action)}$
\EndIf
\State $\text{Collect }D_i = \{(s_{\pi_i}, \pi_{target}(s_{\pi_i})) \}\text{ dataset of visited states by } \pi_i \text{ and actions by expert/coach}$
\State $\text{Aggregate datasets } D \gets D \cup D_i$
\State $\text{Train policy } \pi_{i+1} \text{ on }D$
\EndFor \\
\Return $\text{best } \pi_i \text{ on validation set}$
\end{algorithmic}
\end{algorithm}

\subsubsection{DAgger by coaching}
With DAgger, the policy space can be far from the learning policy space and so it limits the learning ability and information might be not inferable from the state. To prevent this, HHH Daumé III et al. proposed, in 2012, the DAgger by coaching algorithm[6] . With this algorithm, we execute easy-to-learn actions i.e within learner's ability. When it's too hard, the coach lowers the goal and teaches gradually.\\

We define a \textit{hope action}, easier to achieve than the oracle action and which is not so much worse. Let $\lambda_i$ measure how close the coach is to the oracle : $\widetilde{\pi_i} = \arg \max_{a \in A}\ \lambda_i \text{score}_{\pi_i}(s,a) - C(s,a)$. DAgger by coaching guarantees linear regret.

\begin{theorem} Let $\Tilde{l}_i(\pi) = \mathbb{E}_{s \sim d_{\pi_i}}[l(s, \pi, \Tilde{\pi}(s))]$ denote the expected surrogate loss w.r.t. $\widetilde{\pi}$ and $\Tilde{\epsilon}_N = \frac{1}{N} min _{\pi \in \Pi} \sum_{i=1}^N \Tilde{l}_i(\pi)$ denote the true loss of the best policy in hindsight with respect to hope actions, then there exists $\pi \in \pi_{1:N}$ and $u$ such as $J(\pi) \leq J(\pi^*) + u T \Tilde{\epsilon}_N + O(1)$
\end{theorem}

The DAgger algorithm and its equivalent with coaching are described in Algorithm~\ref{alg:dagger}.

%doublefig
\begin{figure}[!htb]
\centering
\subfloat[Original DAgger]         {\includegraphics[width=.25\linewidth]{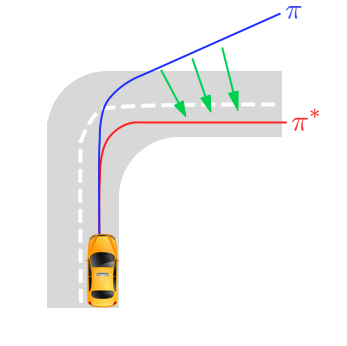}}\hspace{0.5cm}
\subfloat[DAgger by Coaching]  
  {\includegraphics[width=.25\linewidth]{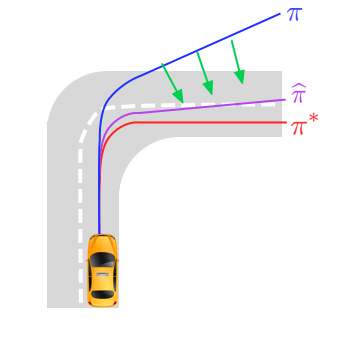}}
\caption{DAggers algorithms illustration on self-driving cars}
\label{fig:dag}
\end{figure}

\begin{figure}[!h]
\centering
\includegraphics[width=.5\linewidth]{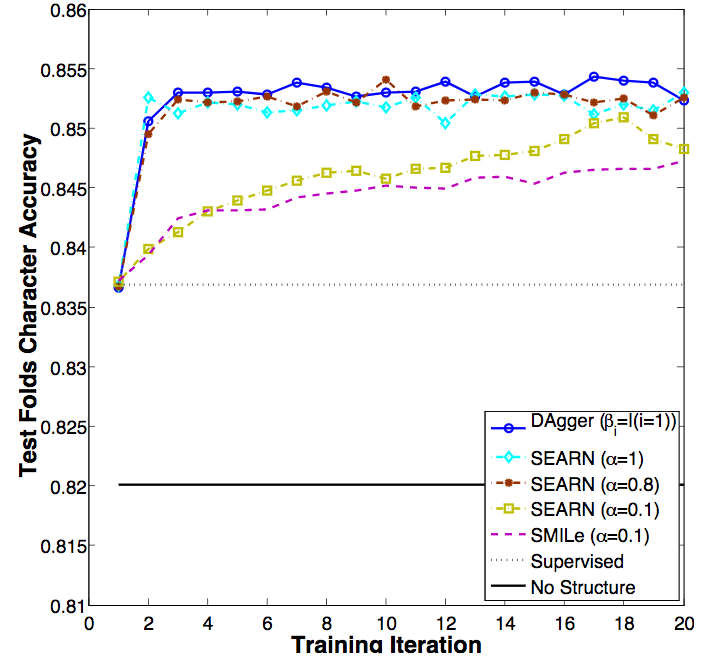}
\caption{Performance comparison between DAgger, SMILe, SEARN and the supervised approach on handwritten character recognition task . The baseline is a just a SVM that predicts each character independently. SEARN with $\alpha = 1$ (equivalent to a pure policy iteration approach), $\alpha = 0.8$ and DAgger that performs better on this task \textit{(source : original DAgger paper [5] )}.}
\label{fig:com}
\end{figure}

\subsection{Approximate Policy Iteration with Demonstration (APID)}
For the previous algorithms, we assumed that the expert exhibits optimal behaviour and that his demonstrations are abundant. Those assumptions are not always valid in the real world and so, in order to address this challenging issue, we combine both expert and interaction data (i.e., mix LfD and RL). APID (2013) [7] is thus particularly interesting in cases where the expert demonstrations are few or suboptimal. It is a LfD with a regularized Approximate Policy Iteration (API) method, the key idea being that expert’s suggestions are used to define linear constraints which guide the optimization performed by API.

Formally, we place ourselves in the context of the API and use the added information furnished by the experts (even if too few or inaccurate). $V^{\pi}$ and $Q^{\pi}$ denote the value and action-value function for $\pi$, and $V^*$ and $Q^*$ denote their corresponding for the optimal policy $\pi^*$. We have a set of interaction data $\mathcal{D}_{RL} = {(S_i,A_i)}_{i=1}^n$, respectively a set of expert examples $\mathcal{D}_{E} = {(S_i,A*_i)}_{i=1}^m$, representing a n examples sample of (state, action) couples, respectively a m examples sample of (state, demonstrated action) couples. In order to encode the suboptimality of the expert, we add a variable to the action-value optimal policy to allow occasional violations of the constraints. Finally, we get a constrained optimization problem. In this approach, we do not have access to the exact Bellman operator $T^\pi$ but only to samples and we thus use the Projected Bellman error.

\subsection{Aggregate Values to Imitate (AggreVaTe)}
%Leverages cost-to-go information. 
AggreVaTe, introduced by Ross and Bagnell (2014) [8], is an extension of the DAgger algorithm that learns to choose actions in order to minimize the cost-to-go (total cost) of the expert rather than the zero-one classification loss of mimicking its actions. For the first iteration, we passively collect data by observing the expert performing the task. In each trajectory, at a uniformly random time $t$, we explore an action $a$ in state $s$ and observe the cost-to-go $Q$ of the expert after performing this action.

We use $Q_t^{\pi}(s,a)$ to denote the expected future cost-to-go of executing action $a$ in state $s$, followed by executing policy $\pi$ for $t-1$ steps.

Exactly as the DAgger algorithm, AggreVaTe collects data through interaction with the learner as : 
\begin{itemize}
    \item At each iteration, we use the current learner policy $\pi_i$ to perform the task, interrupt at a uniformly random time $t$, explore an action a in the current state s, after which control is provided back to the expert to continue up to time-horizon T
    \item  It results in new examples of the cost-to-go of the expert $(s, t, a, Q)$, under the distribution of states visited by the current policy $\pi_i$.
    \item Then we aggregate datasets and train $\pi_{i+1}$on the concatenated datasets
\end{itemize}

The full algorithm is described in Algorithm~\ref{alg:aggrevate}.
%$$\mathbb{E}_{\tau \sim \pi}[\sum_t l_t] < \mathbb{E}_{\tau \sim \pi^*}[\sum_t l_t] +T(\epsilon_{class} + \epsilon_{regret}) + O(\frac{TlogTQ_{max}}{\alpha N})$$
\begin{theorem}Let $\epsilon_{regret} = \frac{1}{N}[\sum_{i=1}^N l_i(\widehat{\pi_i}) - min_{\pi \in \Pi}\sum_{i=1}^N l_i(\pi)]$ denote the online learning average regret and $\epsilon_{class} = min_{\pi \in \Pi}\frac{1}{N}\sum_{i=1}^N \mathbb{E}_{t \sim U(1:T), s \sim d_{\pi_i}^t}[Q^*_{T-t+1}(s,a) - \min_a Q^*_{T-t+1}(s,a)]$ denote the minimum expected cost-sensitive classification regret. Then, there exists $\pi \in \pi_{1:N}$ such as 
$J(\pi) < J(\pi^*) +T(\epsilon_{class} + \epsilon_{regret}) + O(\frac{TlogTQ_{max}}{\alpha N})$
\end{theorem}

AggreVaTe can be interpreted as a regret reduction of imitation learning to no-regret online learning.

\begin{algorithm}
\caption{AggreVaTe}\label{alg:aggrevate}
\begin{algorithmic}[1]
\State $\text{Initialize } D \gets \emptyset \text{, } \widehat{\pi}_1 \gets \text{ any policy in } \Pi$
\For {$i = 1:N$}
\State $\text{Let } \pi_i = \beta_i \pi^* + (1 - \beta_i) \widehat{\pi_i}$
\State $\text{Collect m data points as follows :}$
\For {$j = 1:m$}
\State $\text{Sample uniformly } t \in \{1, 2, ..., T\}$
\State $\text{Start new trajectory in some initial state drawn from initial state distribution}$
\State $\text{Execute current policy } \pi_i \text{ up to time } t-1$
\State $\text{Execute some exploration action } a_t \text{ in current state } s_t \text{ at time } t$
\State $\text{Execute expert from time } t+1 \text{ to } T \text{ and observe estimate of cost-to-go } \widehat{Q} \text{ starting at time } t$
\EndFor
\State $\text{Collect } \mathcal{D}_i = \{(s, t, a, \widehat{Q})\}\text{ dataset of  states, times, actions and expert's cost-to-go}$
\State $\text{Aggregate datasets } \mathcal{D} \gets \mathcal{D} \cup \mathcal{D}_i $
\State $\text{Train cost-sensitive classifier } \widehat{\pi}_{i+1} \text{ on }\mathcal{D}$
\EndFor \\
\Return $\text{best } \widehat{\pi}_i \text{ on validation set}$
\end{algorithmic}
\end{algorithm}

\subsection{Extensions}
We let for future work the literature review of some of the most recent, exciting and promising work in Imitation Learning. Indeed, OpenAI recently proposed a meta-learning framework [9] to achieve imitation learning with a very few expert data. Their goal was to teach a physical robot to stack small color blocks as a child would do. The expert data was given using VR and computer vision and the robot has learnt his stacking task using only one demonstration from an arbitrary situation. The videos are available on their \href{https://blog.openai.com/robots-that-learn/}{\textit{website}}. They achieved it by pretraining their meta-framework on numerous set of tasks using Neural Networks. To train the policy, they mainly used the DAgger algorithm. Then, the robot received one demonstration of an unobserved task and mimicked it. \\
OpenAI (Ho et Ermon) also proposed an Imitation Learning approach [10] based on Generative Adversarial Network (GAN) which aims at training to learn to mimic expert’s demonstrations without an explicit reward. One of the key parts is that their method is model-free and also it does not query the expert during learning. Their approach explores randomly how to determine which actions lead to a policy that best mimics the expert behaviour.

\section{Experiments}
To experiment with imitation learning and especially with the DAgger algorithm, we followed the instructions of a Deep Reinforcement assignment from Berkeley University [11] where the expert demonstrations have already been trained using the OpenAI Gym toolkit and the classifier is a Neural Network trained using TensorFlow. 

The goal is to teach a virtual half cheetah to run and leap, in a straight-forward way. The learner \textit{(our virtual cheetah)} is sequentially asking information (the input - i.e. expert data - is an environment-specific $17 \times 1$ pixel array representing the observation of the said environment) to the expert, then re-training and re-asking when needed. 

\begin{figure}[!htb]
\centering
\subfloat[Expert policy $\pi^*$]      {\includegraphics[width=.25\linewidth]{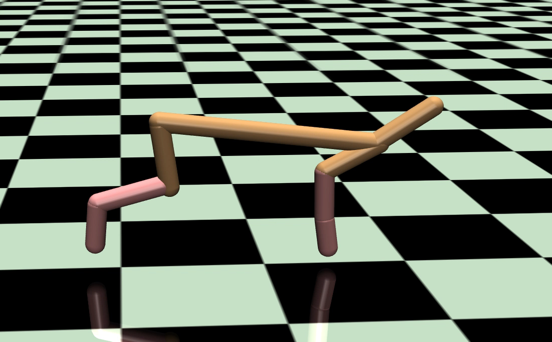} \label{fig:exp}  }\hspace{0.5cm}
\subfloat[First learned policy $\pi_1$]    
{\includegraphics[width=.25\linewidth]{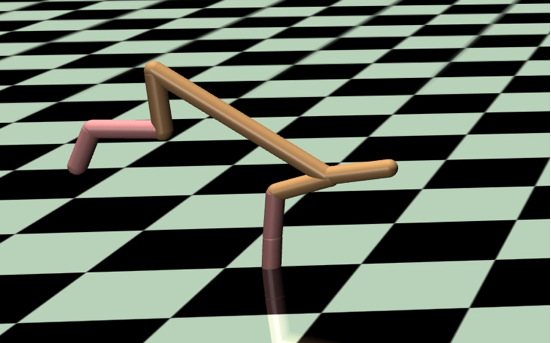}\label{fig:first_l}}\hspace{0.5cm}
\subfloat[Third learned policy $\pi_3$]   
{\includegraphics[width=.25\linewidth]{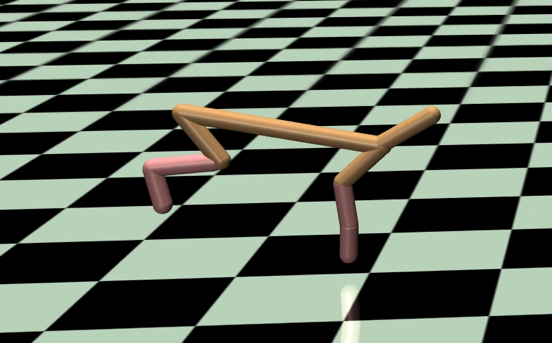}\label{fig:third_l}}
\caption{Policies of the virtual running cheetah when using the OpenAI Gym toolkit, visualizations of the leap landing improvements over policy iterations}
\label{fig:dag}
\end{figure}

We trained a first policy for the expert data (Figure~\ref{fig:exp}), then we runned the first learning policy (Figure~\ref{fig:first_l}) to get a first dataset. Afterwards, we queried the expert to label the dataset with actions and then we aggregated the dataset. The learner is running forward but its leap landings are approximate at the first iteration. However, it keeps improving itself (Figure~\ref{fig:third_l}) and eventually, in the end, we keep the best policy chosen on the validation set. 

\begin{figure}[!h]
\centering
\includegraphics[width=.5\linewidth]{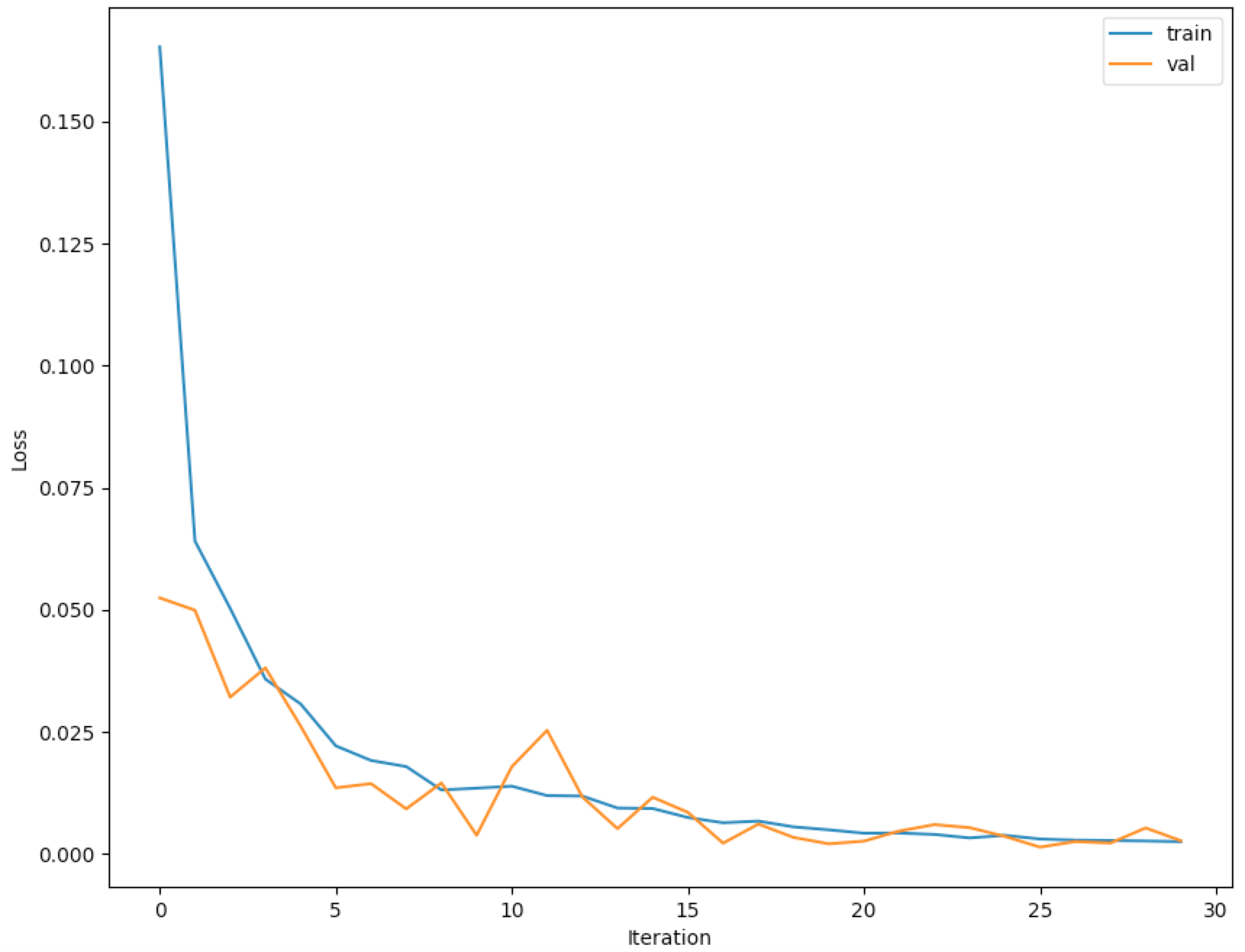}
\caption{Training \textit{(blue curve)} and validation \textit{(orange curve)} losses over iterations during the DAgger training on the cheetah run}
\label{fig:loss}
\end{figure}

The more rollouts as training data, the better the results and we notice that the loss is converging after less than 30 iterations as depicted in Figure~\ref{fig:loss}.

\section{Conclusion}
To conclude, let us recall that the goal of the project was to deeply understand the Imitation Learning tasks, review the main algorithms and their regret bounds to achieve these tasks and then compare them. The performance of the different algorithms depends on the tasks and the nature of the input data. Indeed, the expert policy can be restricted or sometimes inaccurate, it might be too expensive to produce full trajectories, and so on and so force. However, DAgger is nowadays the most common algorithm because it generally outperforms the other models. Hence, we applied it using the OpenAI Gym toolkit to measure its performance and analyze the learner progression during the said process. We let, for future work, the comparison with other Imitation Learning approaches using this toolkit. \\
In short, this project was both enriching and exciting. We had the chance to review the different approaches to perform imitation learning and explore the new applications of this wide field.

% References without cites
%\nocite{sparse_auto}
\bibliographystyle{unsrt}
\bibliography{bibliography}
\section*{References} \fontsize{8.5}{1.5pt}\selectfont
[1] Pieter Abbeel and Andrew Y. Ng. Apprenticeship learning via inverse reinforcement learning. In Proceedings of the Twenty-first International Conference on Machine Learning, ICML ?04, pages 1?, New York, NY, USA, 2004. ACM.

[2] Stephane Ross and Drew Bagnell. Efficient reductions for imitation learning. In Yee Whye Teh and Mike Titterington, editors, Proceedings of the Thirteenth International Conference on Artificial Intelligence and Statistics, volume 9 of Proceedings of Machine Learning Research, pages 661?668, Chia Laguna Resort, Sardinia, Italy, 13?15 May 2010. PMLR.

[3] Hal Daum\'e III, John Langford, and Daniel Marcu. Search-based structured prediction. CoRR, abs/0907.0786, 2009.

[4] Kshitij Judah, Alan Fern, and Thomas G. Dietterich. Active imitation learning via reduction to I.I.D. active learning. CoRR, abs/1210.4876, 2012.

[5] St\'ephane Ross, Geoffrey J. Gordon, and J. Andrew Bagnell. No-regret reductions for imitation learning and structured prediction. CoRR, abs/1011.0686, 2010.

[6] He He, Hal Daum\'e, III, and Jason Eisner. Imitation learning by coaching. In Proceedings of the 25th International Conference on Neural Information Processing Systems - Volume 2, NIPS?12, pages 3149?3157, USA, 2012. Curran Associates Inc.

[7] Beomjoon Kim, Amir-massoud Farahmand, Joelle Pineau, and Doina Precup. Learning from limited demonstrations. In Proceedings of the 26th International Conference on Neural In- formation Processing Systems - Volume 2, NIPS?13, pages 2859?2867, USA, 2013. Curran Associates Inc.

[8] St\'ephane Ross and J. Andrew Bagnell. Reinforcement and imitation learning via interactive no-regret learning. CoRR, abs/1406.5979, 2014.

[9] Yan Duan, Marcin Andrychowicz, Bradly C. Stadie, Jonathan Ho, Jonas Schneider, Ilya Sutskever, Pieter Abbeel, and Wojciech Zaremba. One-shot imitation learning. CoRR, abs/1703.07326, 2017.

[10] Jonathan Ho and Stefano Ermon. Generative adversarial imitation learning. CoRR, abs/1606.03476, 2016.

[11] UC Berkeley CS 294: Deep RL assignment 1: Imitation learning. \url{http://rll.berkeley. edu/deeprlcoursesp17/docs/hw1.pdf}. Accessed: 2018-01-14.

\newpage
\appendix
\normalsize\section{Appendix}\label{sec:appendix}
Please find, in the following section, proofs (entire proof, extracts or intuitions) of the theorems detailed above, there were proposed by their respective papers :
\subsection{Proof Theorem 1 (Supervised Learning)}
Let $\epsilon_i=\mathbb{E}_{s \sim d^i_{\pi^*}}[e_{\widehat{\pi}}(s)] $ for $i=1,2,...,T$ the expected 0-1 loss at time $i$ of $\widehat{\pi}$, such that $\epsilon=\frac{1}{T}\sum_i \epsilon_i$.   Note that $\epsilon_t$ corresponds to the probability that $\widehat{\pi}$ makes a mistake under  distribution $d^t_{\pi^*}$. Let $p_t$ represent  the probability $\widehat{\pi}$ has not  made a  mistake  (w.r.t. $\pi^*$) in  the  first $t$-step, and $d_t$ the distribution of state $\widehat{\pi}$ is in at time $t$ conditioned on the fact it hasn’t made a mistake so far. \\
If $d'_t$ represents the distribution of states at time $t$ obtained by following $\pi^*$ but  conditioned  on  the  fact  that $\widehat{\pi}$ made  at least  one  mistake  in  the  first $t-1$ visited  states.   Then, $$d^t_{\pi^*}=p_{t-1}d_t+ (1-p_{t-1})d'_t$$ 
Now  at  time $t$,  the  expected cost of $\widehat{\pi}$ is at most 1 if it has made a mistake so far, or $\mathbb{E}_{s \sim d_t}(C_{\widehat{\pi}}(s))$ if it hasn’t made a mistake yet.  So $$J(\widehat{\pi}) \leq \sum _t[p_{t-1}\mathbb{E}_{s\sim d_t}(C_{\widehat{\pi}}(s))+(1-p_{t-1})]$$ 
Let $e_t$ and $e'_t$ represent the probability of mistake of $\widehat{\pi}$ in distribution $d_t$ and $d'_t$. Then $$\mathbb{E}_{s \sim d_t}(C_{\widehat{\pi}}(s)) \leq \mathbb{E}_{s\sim d_t}(C_{\pi^*}(s)) + e_t$$ and since $\epsilon_t=p_{t-1}e_t + (1-p_{t-1})e'_t$, then $p_{t-1}e_t \leq \epsilon_t$.\\ 
Additionally since $p_t= (1-e_t) p_{t-1}$, $p_t \geq p_{t-1}-\epsilon_t \geq 1-\sum_{i=1}^t \epsilon_i$, i.e. $1-p_t \leq \sum_i \epsilon_i$. Finally note that $$J(\pi^*) = \sum_{t=1}^T[p_{t-1}\mathbb{E}_{s \sim d_t}(C_{\pi^*}(s))+(1-p_{t-1})\mathbb{E}_{s \sim d'_t}(C_{\pi^*}(s))]$$ so that $$\sum_t p_{t-1}\mathbb{E}_{s \sim d_t}(C_{\pi^*}(s)) \leq J(\pi^*)$$ Using these facts we obtain:
\begin{align*}
    J(\widehat{\pi}) \leq {}& \sum_{t=1}^T[p_{t-1}\mathbb{E}_{s \sim d_t}(C_{\widehat{\pi}}(s)) + (1-p_{t-1})]\\
     \leq {}& \sum_{t=1}^T[p_{t-1}\mathbb{E}_{s \sim d_t}(C_{\pi^*}(s)) + p_{t-1}e_t+(1-p_{t-1})]\\
     \leq {}& J(\pi^*) + \sum_{t=1}^T \sum_{i=1}^t \epsilon_i\\
     \leq {}& J(\pi^*) + T \sum_{t=1}^T \epsilon_t\\
     = {}& J(\pi^*) + T^2 \epsilon
\end{align*}

\subsection{Proof Theorem 2 (Forward Training)}
We follow here a similar proof than the previous one. We denote $Q^{\pi'}_t(s, \pi)$ the $t$-step cost of executing $\pi$ in initial state s and then following policy $\pi'$. Let $Q^{\pi^*}_{T-t+1}(s, a) - Q^{\pi^*}_{T-t+1}(s, \pi^*) < u$ and assume that $l(s, \pi)$ is an upper bound on the $0-1$ loss. At iteration $i$ we are only changing the policy at step $i$, so 
$$ J(\pi^i) =  J^{\pi^{i-1}}(\pi^i_i,i) = J(\pi^{i-1}) + [J(\pi_{1:T-i} - J(\pi_{1:T-i-1})]$$
Solving this recurrence proves :
\begin{align*}
    J(\pi) = {}& J(\pi^*) + \sum_{i=0}^{T-1} [J(\pi_{1:T-i} - J(\pi_{1:T-i-1})]\\
     = {}& J(\pi^*) + \sum_{i=1}^T \mathbb{E}_{s \sim d_{\pi}^i}[Q^{\pi^*}_{T-i+1}(s, \pi) - Q^{\pi^*}_{T-i+1}(s, \pi^*)]\\
     \leq {}& J(\pi^*) + u \sum_{i=1}^T \mathbb{E}_{s \sim d_{\pi}^i}[l(s, \pi)]\\
    = {}& J(\pi^*) + u T \epsilon
\end{align*}

\subsection{Proof Theorem 3 (SEARN)}
Let define $\Bar{J}_k^{\pi}(\pi')$ the expected $T$-step cost of executing $\pi'$ $k$ times and policy $\pi$ at all other steps. The SEARN algorithm seeks to minimize directly the bound :
$$J(\pi^n) \leq J(\pi^*) + T \alpha (1-\alpha)^{T-1}\sum_{j=1}^n (\Bar{J}_1^{\pi^{j-1}} (\widehat{\pi}^{j}) - J(\pi^{j-1})) + n \alpha^2 \frac{T^2(T-1)}{2}$$
by choosing $\hat{\pi}^n$ to minimize $\Bar{J}_1^{\pi^{n-1}} (\widehat{\pi}^{n}) - J(\pi^{n-1})$. Using $N=\frac{2}{\alpha}lnT$, $\alpha = \frac{1}{T^3}$ and denoting $\mathbb{A}_1 = \frac{1}{n}\sum_{j=1}^n (\Bar{J}_1^{\pi^{j-1}} (\widehat{\pi}^{j}) - J(\pi^{j-1}))$, SEARN guarantees :
$$J(\pi^n) \leq J(\pi^*) + O(TlnT\mathbb{A}_1 + lnT)$$
For  each  state, the cost-to-go under the current policy must be estimated for each action during a cost-sensitive classification problem. 

\subsection{Proof Theorem 4 (SMILe)}
Since for SMILe, $\widehat{\pi}^{n+1}$ will be close to $\pi^n$, we can derive bounds on the policy disadvantages. Let $J_k^{\pi}(\pi')$ denote the expected $T$-step cost of executing $\pi'$ at steps $\{t_1, ..., t_k\}$ and $\Bar{J}_k^{\pi}(\pi')$ the expected $T$-step cost of executing $\pi'$ $k$ times and policy $\pi$ at all other steps. The bound follows from the fact when $\widehat{\pi}^{n+1}$ acts like $\pi^*$ at time step $t$ :
$$\Bar{J}_2^{\pi^n}(\widehat{\pi}^{n+1}) - J(\pi^n) = 2[\Bar{J}_1^{\pi^n}(\widehat{\pi}^{n+1}) - J(\pi^n)]$$
Moreover, if $\alpha < 1/T$, then \textit{(Lemma 4.1 in the Ross and Bagnel \cite{SMILe})}:
\begin{align*}
\begin{split}
 J(\pi^n) \leq {}& J(\pi^*) + [\alpha T (1-\alpha)^{T-1}\sum_{j=1}^n(\Bar{J}_1^{\pi^{j-1}} (\widehat{\pi}^{j}) - J(\pi^{j-1})) 
 \\& + \alpha^2 \frac{T(T-1)}{2} (1-\alpha)^{T-2}\sum_{j=1}^n(\Bar{J}_2^{\pi^{j-1}} (\widehat{\pi}^{j}) - J(\pi^{j-1}))
  \\& + n \alpha^3 T {T\choose 3}
\end{split}
\end{align*}
and if $n > \frac{2}{\alpha} ln T$ \textit{(Lemma 4.2 in the Ross and Bagnel \cite{SMILe})}:
$$J(\Tilde{\pi}^n) \leq J(\pi^n) + 1$$
By denoting $\mathbb{A}_1= \sum_{j=1}^n(\Bar{J}_1^{\pi^{j-1}} (\widehat{\pi}^{j}) - J(\pi^{j-1}))$ and\\ $\Tilde{\epsilon} = \frac{\alpha}{1-(1-\alpha)^n}\sum_{i=1}^{n}(1-\alpha)^{i-1} \mathbb{E}_{s \sim d_{\pi^{i-1}} }(e(s,\widehat{\pi}^{*i}))$ it follows that with $\alpha$ in $O(\frac{1}{T^2})$ and $n$ in \\$O(T^2 ln T)$ :
$$ J(\pi^n) \leq J(\pi^*) + O(T(\mathbb{A}_1 + \Tilde{\epsilon}) + 1)$$
 
\subsection{Proof Theorem 5 (DAgger)}
Let $\epsilon_N = min _{\pi \in \Pi} \frac{1}{N}\sum_{i=1}^N \mathbb{E}_{s \sim d_{\pi_i}}[l(s,\pi)]$ be the true loss of the best policy in hindsight then if $N$ is $O(T)$ there exists a policy $\widehat{\pi} \in \widehat{\pi_{1:N}}$ such that 
$$\mathbb{E}_{s \sim d_{\widehat{\pi}}}[l(s,\widehat{\pi})] \leq \epsilon_N + O(1/T) $$
For an arbitrary task cost function $C$, if $l$ is an upper bound on the $0-1$ loss with respect to $\pi^*$, combining this results with Theorem 2.2 (Forward Training) yields that if $N$ is $O(uT)$ there exists a policy $\widehat{\pi} \in \widehat{\pi_{1:N}}$ such that
$$J(\widehat{\pi}) \leq J(\pi^*) + u T \epsilon_N + O(1)$$

\subsection{Proof Theorem 6 (DAgger by coaching)}

Similar proof than for the theorem 5 by first deriving a regret bound for coaching.
\end{document}